\documentclass[10pt, conference, compsocconf]{IEEEtran}
\ifCLASSINFOpdf
\else
\fi
\usepackage{graphicx}  
\usepackage{amsmath}
\usepackage{caption}
\usepackage{booktabs}
\usepackage{multirow}
\usepackage{dingbat}
\usepackage{subfig}
\usepackage{enumerate}
\usepackage{amsmath}
\usepackage{array}

\begin{document}
%
\title{NRTR: A No-Recurrence Sequence-to-Sequence Model For Scene Text Recognition}


\author{\IEEEauthorblockN{Fenfen Sheng}
\IEEEauthorblockA{Institute of Automation, Chinese Academy of Sciences\\
University of Chinese Academy of Sciences\\
Beijing 100190, China\\
shengfenfen2015@ia.ac.cn}
\and
\IEEEauthorblockN{Zhineng Chen}
\IEEEauthorblockA{Institute of Automation, Chinese Academy of Sciences\\
Beijing 100190, China\\
zhineng.chen@ia.ac.cn}
\and
\IEEEauthorblockN{Bo Xu}
\IEEEauthorblockA{Institute of Automation, Chinese Academy of Sciences\\
Beijing 100190, China\\
xubo@ia.ac.cn}
}


%


\maketitle

\begin{abstract}
Scene text recognition has attracted a great many researches due to its importance to various applications.
Existing methods mainly adopt recurrence or convolution based networks.
Though have obtained good performance, these methods still suffer from two limitations: slow training speed due to the internal recurrence of RNNs, and high complexity due to stacked convolutional layers for long-term feature extraction.
This paper, for the first time, proposes a no-recurrence sequence-to-sequence text recognizer, named NRTR, that dispenses with recurrences and convolutions entirely.
NRTR follows the encoder-decoder paradigm, where the encoder uses stacked self-attention to extract image features, and the decoder applies stacked self-attention to recognize texts based on encoder output.
NRTR relies solely on self-attention mechanism thus could be trained with more parallelization and less complexity.
Considering scene image has large variation in text and background, we further design a modality-transform block to effectively transform 2D input images to 1D sequences, combined with the encoder to extract more discriminative features.
NRTR achieves state-of-the-art or highly competitive performance on both regular and irregular benchmarks, while requires only a small fraction of training time compared to the best model from the literature (at least $8$ times faster).
\end{abstract}


%
\IEEEpeerreviewmaketitle

\section{Introduction}
\begin{figure}
\footnotesize
\centering
\includegraphics[width=8cm]{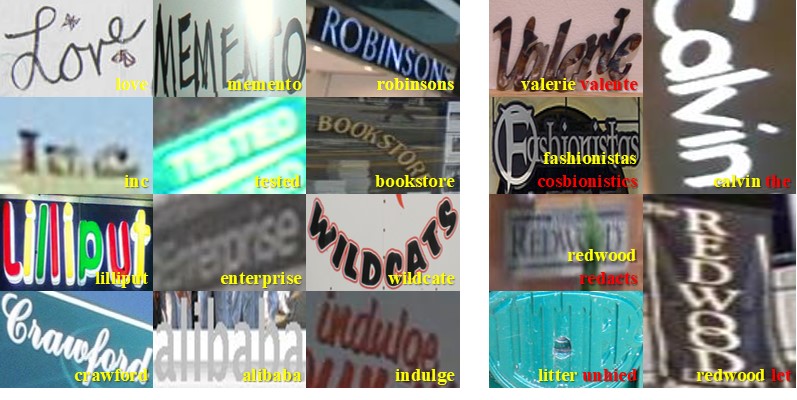}
\captionsetup{font={footnotesize}}
\caption{Qualitative results of NRTR. (left) Correct results with texts in yellow. (right) Incorrect results with labels in yellow and outputs in red.}
\label{fig4}
\end{figure}

Scene text recognition has drawn increasing interests as it could extract rich semantic information relevant to scene and object. Although extensive studies have been carried out, recognizing scene texts is still challenging due to its high complexity, e.g., low quality texts, arbitrary orientations, cluttered backgrounds and complex deformations, see Fig.\ref{fig4}.

Current text recognition methods \cite{bai2018edit,cheng2017focusing,shi2017end,gao2017reading,yin2017scene,shi2016robust} mainly follow the sequence-to-sequence (seq2seq) paradigm, where input images and output texts are separately represented as patch sequences and character sequences. These methods could be roughly classified into two branches: the recurrent neural network (RNN) based recognizers and the convolutional neural network (CNN) based ones.

RNN based text recognizers \cite{bai2018edit,shi2017end,cheng2017focusing,shi2016robust,he2016reading} have show great success as they are superior to learn contextual information and capture strong correlation among different characters in each text.
However, the inherently sequential nature of RNN precludes computation parallelization, which brings heavy time and computational burdens when input image sequence is long, as memory constraints limit batching across examples.
Besides, the training procedure of RNN is sometimes tricky due to the problem of gradient vanishing/exploding \cite{bengio1994learning}.

Recently, CNN based recognizers \cite{gao2017reading,yin2017scene} are proposed to accelerate sequential computation.
By leveraging CNN instead of RNN, they enable to compute hidden representation in parallel.
However, the number of operations required to relate two arbitrary signals grows along with their distances. CNN based methods are difficult to learn dependencies among distant positions, unless much more convolutional layers are stacked, which in turn increases the complexity.
Therefore, CNN based methods suffer from the dilemma of low complexity and satisfactory performance.


In this paper, we propose, for the first time, a no-recurrence seq2seq scene text recognizer, named NRTR, that dispenses with recurrences and convolutions entirely. Motivated by recent success of Transformer \cite{vaswani2017attention} in natural language processing field, NRTR relies solely on self-attention mechanism.
Specifically, NRTR follows the encoder-decoder framework, while the encoder uses stacked self-attention to transform input image sequence to hidden feature representation, and the decoder applies stacked self-attention to output sequence of characters based on the encoder output.
NRTR draws global dependencies between different input and output positions at once rather than one by one in RNN, and reduces the whole operation to a constant number unlike that in CNN.
It therefore allows for more computation parallelization with higher performance.

Besides, unlike \cite{vaswani2017attention} that uses 1D sentence as input, scene text recognizer receives 2D images with large variation in scales/aspect ratios and backgrounds.
We further proposes a novel modality-transform block as a preprocessing step before the encoder, to effectively convert 2D image to corresponding 1D sequence.
Experiments demonstrate that the specially designed modality-transform block could greatly influence the whole recognizer performance.

\begin{figure*}
\centering
\includegraphics[width=13cm, height=6cm]{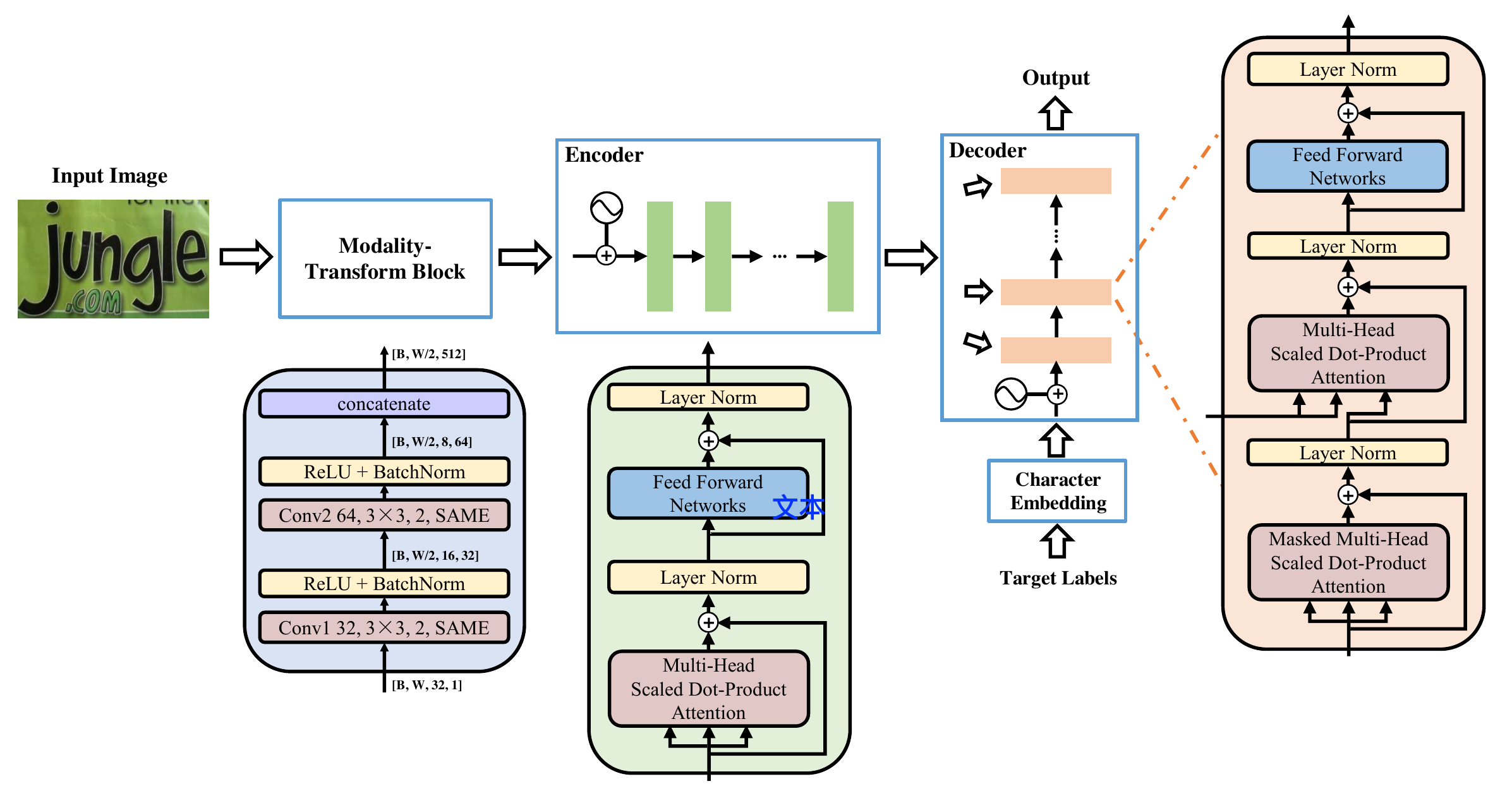}
\caption{The overall architecture of NRTR.}
\label{fig2}
\end{figure*}

We conduct extensive experiments on standard benchmarks, including both regular (IIIT5K, SVT, ICDAR2003 and ICDAR2013) and irregular datasets (SVT-P, CUTE80, ICDAR2015).
Without the bells and whistles, NRTR achieves state-of-the-art or highly competitive performance in both lexicon-free and lexicon-based cases without any rectification module, while accompanies with a $8$ times faster training speed than the existing best recognizer.

Our contributions are summarized as follows:
\begin{itemize}
\item We analysis, for the first time, the deficiency of current RNN and CNN based recognizers and propose a no-recurrence seq2seq model based solely on self-attention mechanism. Our model could be trained with more computation parallelization and less complexity.
\item We design a modality-transform block to efficiently map input image to corresponding sequence, combined with encoder, to extract more discriminative features.
\item Our model achieves state-of-the-art performance on various benchmarks and significantly surpasses competitive recognizers on both accuracy and training speed.
\end{itemize}

\section{Related Work}
\subsection{Scene Text Recognition}
Traditional text recognizers mainly adopt bottom-up scheme by first detecting individual characters using sliding window \cite{wang2011end}, then integrating characters into texts by dynamic programming or lexicon search \cite{wang2011end}. Others adopt top-down scheme by directly recognizing texts from images.

Recent methods regard text recognition as a sequence recognition problem where images and texts are represented as patch and character sequences separately.
Shi et al. \cite{shi2017end} combine CNN and RNN to learn spatial dependencies and applies CTC to translate per-slice prediction into a label sequence. They also develop an attention-based spatial transformer network to rectify irregular texts \cite{shi2016robust}. Besides, Lee et al. \cite{lee2016recursive} and Cheng et al. \cite{cheng2017focusing} both construct attention-based recurrent network to decode feature sequence and predict labels recurrently. Instead of RNNs, Gao et al. \cite{gao2017reading} and Yin et al. \cite{yin2017scene} leverage stacked CNN for the pursuit of greater computational parallelism.

\subsection{Our Method Versus Some Related Works}
The most related work to NRTR is Transformer, a recent development in NLP field.
Inspired by Transformer, we uses solely self-attention mechanism as the fundamental module but has distinct differences indeed.
First, Transformer aims for machine-based English-to-French translation task but fails to read texts in natural images. As input text images generally convey much more information than sentences in machine translation, text recognizers tend to be more complicated. Our model relies on the specifically designed encoder/decoder to efficiently solve this problem.
Second, unlike Transformer that receives 1D sentences as input, we use 2D images with large variation in scales/aspect ratios and backgrounds. The locations and geometric features of scene texts lying in images play a more complex role than the word location in an 1D sentence. To obtain the most useful sequence for the encoder, we design a novel modality-transform block to transform each input image effectively.

\section{Methodology}
The architecture of NRTR is depicted in Fig.\ref{fig2}. NRTR consists of three sub-networks: the encoder, the decoder and the modality-transform block served as the preprocessing.
As the encoder and the decoder are both based on the self-attention mechanism, we first review it and then describe the three main sub-networks.

\subsection{Self-Attention Mechanism}
Self-attention extracts correlation information between different input and output positions. Here, we use Scaled Dot-Product Attention, an effective self-attention module proposed in \cite{vaswani2017attention}. It has three inputs: queries and keys of dimension ${d_k}$, and values of dimension ${d_v}$. Dot product is performed between the query and all keys to obtain their similarity. A softmax function is applied to obtain the weights on the values. Given a query $\bf{q}$, all keys (packed into matrices $\bf{K}$) and values (packed into $\bf{V}$), the output value is weighted average over input values:
\begin{equation}
{v^{out}} = {\mathop{\rm softmax}\nolimits} \left( {{{{\bf{q}}{{\bf{K}}^t}} \over {\sqrt {{d_k}} }}} \right){\bf{V}}
\end{equation}
where ${t}$ means element numbers of corresponding inputs and scalar ${1 \over {\sqrt {{d_k}} }}$ is used to prevent softmax into regions where it has extremely small gradients. Therefore, self-attention could connect all positions with a constant number and allow parallelization. More details please refer to \cite{vaswani2017attention}.

\subsection{Modality-Transform Block}
Modality-transform block consists of several convolutional layers. For each layer, the stride is set to $2$ and channel number is increased progressively by $2\times$. The product of the height and channel number of each layer therefore remains constant and equal to ${d_{{\rm{model}}}}$, the dimension used in our encoder-decoder model.
More specifically, for each image with $\left( {{w_0},{h_0}} \right)$, at the $n$-th layer, we get
$\left( {w,h,c} \right) = \left( {{\raise0.7ex\hbox{${{w_0}}$} \!\mathord{\left/
 {\vphantom {{{w_0}} {{2^n}}}}\right.\kern-\nulldelimiterspace}
\!\lower0.7ex\hbox{${{2^n}}$}},{\raise0.7ex\hbox{${{h_0}}$} \!\mathord{\left/
 {\vphantom {{{h_0}} {{2^n}}}}\right.\kern-\nulldelimiterspace}
\!\lower0.7ex\hbox{${{2^n}}$}},\left( {{\raise0.7ex\hbox{${d_{{\rm{model}}}}$} \!\mathord{\left/
 {\vphantom {{{d_{model}}} {{h_0}}}}\right.\kern-\nulldelimiterspace}
\!\lower0.7ex\hbox{${{h_0}}$}}} \right) \times {2^n}} \right)$, where $c$ means channel number.
After the final layer, a concatenate operation is applied to reshape features from different channels into an input sequence $\left( {input\_step,dim} \right) = \left( {{\raise0.7ex\hbox{${{w_0}}$} \!\mathord{\left/
 {\vphantom {{{w_0}} {{2^n}}}}\right.\kern-\nulldelimiterspace}
\!\lower0.7ex\hbox{${{2^n}}$}},{d_{{\rm{model}}}}} \right)$, each element of which has ${d_{{\rm{model}}}}$ dimensions. We design various block architectures to observe their influences on the whole model. More information is detailed in the experiment part.

Additionally, as NRTR contains no recurrences, we use positional encoding to indicate each position in sequence:
\begin{equation}
PE_{(pos,i)} = \left\{
\begin{aligned}
sin(pos/10000^{2i/{d_{{\rm{model}}}}}) &\  0 \le i \le {d_{{\rm{model}}}}/2 \\
cos(pos/10000^{2i/{d_{{\rm{model}}}}}) &\  {d_{{\rm{model}}}}/2\le i \le {d_{{\rm{model}}}} \\
\end{aligned}
\right.
\end{equation}
\noindent where ${pos}$ indicates the position in input image sequence and $i$ indicates the $i$-th dimension. We choose this function since for arbitrary fixed offset $k$, ${{PE}_{pos + k}}$ can be represented as a linear function of ${{PE}_{pos}}$. We get the final input sequence by adding the positional encoding to the above input sequence.

\subsection{Encoder}
The encoder consists of ${N_e}$ number of connected identical encoder-blocks (green block in Fig.\ref{fig2}), each of which contains two sub-layers: a multi-head scaled dot-product attention and a position-wise fully connected network.

The multi-head scaled dot-product attention allows the encoder to jointly attend to information from different representation subspaces at different positions. Similar to a convolutional layer that applies a set of filters to extract various features, multi-head attention stacks $h$ times scaled dot-product attention, where $h$ is called the head number. The entire process of the multi-head scaled dot-product attention includes three operations. Firstly, each scaled dot-product attention goes through three different linear projections to project the queries, keys, and values from the input sequence to more discriminative representations. Then, the stacked $h$ times scaled dot-product attentions are performed in parallel, and finally, their outputs are concatenated and undergo a linear layer to get the final ${d_{{\rm{model}}}}$-dimensional outputs:
\begin{equation}
    {\rm{MultiHead}}\left( {{\bf{Q}},{\bf{K}},{\bf{V}}} \right) = {\rm{Concat}}\left( {{\rm{hea}}{{\rm{d}}_{\rm{1}}}{\rm{, \ldots ,hea}}{{\rm{d}}_{\rm{h}}}} \right){{\bf{W}}^O}
\end{equation}
\begin{equation}
    {\rm{where,hea}}{{\rm{d}}_{\rm{i}}} = {\rm{Attention}}\left( {{\bf{QW}}_i^Q,{\bf{KW}}_i^K,{\bf{VW}}_i^V} \right)
\end{equation}
Since ${{\bf{Q,K,V}}}$ have the same dimension of ${d_{{\rm{model}}}}$, the predictions are parameter matrices ${\bf{W}}_i^Q \in {\Re^{{d_{{\rm{model}}}} \times {d_q}}}$, ${\bf{W}}_i^K \in {\Re^{{d_{{\rm{model}}}} \times {d_k}}}$, ${\bf{W}}_i^V \in {\Re^{{d_{{\rm{model}}}} \times {d_v}}}$ and ${\bf{W}}_i^O \in {\Re^{{d_c} \times {d_{{\rm{model}}}}}}$, where ${d_c} = {h \times d_{v}}$, and we set ${d_q} = {d_k} = {d_v} = {d_{{\rm{model}}}}$.

The position-wise fully connected network consists of two linear transformations with a RELU activation in between.
\begin{equation}
    {\rm{FFN}}\left( x \right) = \max \left( {0,x{{\bf{W}}_1} + {b_1}} \right){{\bf{W}}_2} + {b_2}
\end{equation}
\noindent where the weights are ${{\bf{W}}_1} \in {\Re^{{d_{{\rm{model}}}} \times {d_{ff}}}}$ and ${{\bf{W}}_2} \in {\Re^{{d_{ff}} \times {d_{{\rm{model}}}}}}$, and the bias are ${b_1} \in {\Re^{{d_{ff}}}}$ and ${b_2} \in {\Re^{{d_{{\rm{model}}}}}}$. The linear transformations are the same across different positions, but use different parameters from layer to layer.

Meanwhile, layer normalization and residual connection are introduced into each sub-layer for effective training. Given each sub-layer $x$, the corresponding outputs are:
\begin{equation}
    LayerNorm\left( {x + Sublayer\left( x \right)} \right)
\end{equation}

\subsection{Decoder}
The decoder generate text sequence based on encoder outputs and input labels. For each input label, we apply a learnable character-level embedding to convert per character to a $d_{{\rm{model}}}$-dimensional vector. The resulted vectors combines with the positional encoding to form the decoder input.

The decoder consists of ${N_d}$ number of connected identical decoder-blocks (orange block in Fig.\ref{fig2}). Similar to the encoder, the decoder-block is based on the multi-head scaled dot-product attention and the position-wise fully connected network, but has two differences. Firstly, due to the auto-regressive property, a masked multi-head attention is added to each decoder-block to ensure that the predictions for position $j$ can only depend on the known outputs prior to $j$. We implement this by masking out (setting to $-\infty$) all values in the input of the softmax which correspond to illegal connections. Second, the multi-head attention has keys and values coming from the encoder outputs, and queries coming from the previous decoder block outputs.

The outputs are transformed to the probabilities for character classes by a linear projection and a softmax function.

\section{Experiment}
In this section, we describe the standard benchmarks, the detailed experimental settings and results with comparisons.

\subsection{Benchmark Datasets}
\noindent \textbf{IIIT5K} \cite{mishra2012scene} contains 3000 text images in test set. Each is accompanied with a 50-word and a 1k-word lexicons.

\noindent \textbf{SVT} \cite{wang2011end} is collected from Google Street View and most images are severely corrupted by noise and blur. It contains 647 text images, each of which has a 50-word lexicon.

\noindent \textbf{ICDAR 2003 (IC03)} contains 1065 cropped text images after data filtering as in \cite{wang2011end}. Each image is associated with a 50-word and a full lexicon as defined in \cite{wang2011end}.

\noindent \textbf{ICDAR 2013 (IC13)} includes texts on sign boards and objects with large variations. After filtered as done in IC03, it finally contains 1015 cropped text images in test set.

\noindent \textbf{ICDAR 2015 (IC15)} is taken from Google Glasses and contains plenty of irregular texts. For fair comparison, we discard the images that contain non-alphanumeric characters and finally obtain 1922 ones. No lexicon is specified.

\noindent \textbf{SVT-P} \cite{quy2013recognizing} contains 645 cropped text images captured from the side-view angles in Google Street View. Each image is specified with a 50 words lexicon and a full lexicon.

\noindent \textbf{CUTE80} \cite{risnumawan2014robust} is specially collected for evaluating the performance of curved text recognition. It contains 287 cropped text images in test set. No lexicon is provided.

\subsection{Implementation Details}
NRTR is trained purely on Synth90k \cite{jaderberg2014synthetic} and evaluated on standard benchmarks without any finetuning.
In both training and inference, heights of input images are set to $32$ and widths are proportionally scaled.
The output consists of 38 classes, including 26 lowercase letters, 10 digital numbers, 1 space and 1 end-of-sequence token.

At training, samples are batched together by approximate image widths. We use Adam with ${\beta _1} = 0.9$, ${\beta _2} = 0.98$, ${\epsilon}=10^{-9}$, and vary learning rate according to the formula:
\begin{equation}
   lrate = d_{model}^{ - 0.5} \cdot \min \left( {{n^{ - 0.5}},n \cdot warmup\_{n^{ - 1.5}}} \right)
\end{equation}
where $n$ represents current step and $warmup\_n$ (set to $16000$) controls over the learning rate first increase then decrease.
In order to prevent over-fitting, we set residual dropout to $0.1$.
We train NRTR for about 6 epochs before convergence and average the last $10$ checkpoints for inference. All the experiments are implemented in Tensorflow with one Titan X GPU.

\subsection{Ablation Study}
We first investigate the configuration of three sub-networks in NRTR. All experiments are executed with same training strategy and evaluated under the lexicon-free case.

\subsubsection{Exploration of the encoder and the decoder}
We explore the encoder-block number ${N_e}$, the decoder-block number ${N_d}$ and the fully connected inner dimension ${d_{ff}}$. We set ${d_{model}} = 512$, $h = 8$ and the modality-transform block with two convolutional layers during these experiments.
As listed in Tab.\ref{tab:table2}, we take the 6enc6dec model as our baseline.
We first keep total block number identical and find that more encoder blocks achieve better accuracy (8enc4dec vs 6enc6dec, 4enc8dec).
Then, we add more blocks and see that deeper model obtains higher accuracy (12enc6dec vs 8enc4dec, 10enc5dec).
We no longer increase ${N_e}$ and ${N_d}$ considering time and memory costs.
These two comparisons indicate that deeper encoder/decoder could extract more representation, while image information is tend to be more complex than target ones (8enc4dec vs 4enc8dec).
We further test different ${d_{ff}}$ and observe that wider inner dimension is more beneficial to NRTR (6enc6dec-4096 vs 6enc6dec-2048,6enc6dec).
Based on the above analysis, we take the 12enc6dec-4096 as our big model.

\begin{table}
\footnotesize
        \captionsetup{font={footnotesize}}
        \centering
          \scalebox{0.6}{
          \begin{tabular}{l|ccccccc}
            \toprule
            Model                 &${N_e}$  &${N_d}$  &${d_{ff}}$  &IIIT5K &SVT &IC03 &IC13 \\
            \midrule
            6enc6dec (base)       &6        &6        &1024        &85.4      &86.8   &93.5    &92.8 \\
            12enc6dec-4096 (big)  &12       &6        &4096        &86.5      &88.3   &95.4    &94.7 \\
            \midrule
            8enc4dec              &8        &4        &1024        &85.7      &86.6   &93.7    &93.9 \\
            4enc8dec              &4        &8        &1024        &85.2      &86.5   &93.0    &93.2 \\
            10enc5dec             &10       &5        &1024        &85.9      &87.2   &93.9    &94.2 \\
            12enc6dec             &12       &6        &1024        &86.2      &87.7   &94.5    &94.2 \\
            \midrule
            6enc6dec-2048         &6        &6        &2048        &85.9      &87.4   &94.2    &93.9 \\
            6enc6dec-4096         &6        &6        &4096        &86.3      &87.5   &95.1    &93.9 \\
            \midrule
            \midrule
            base model with 2Conv     &6        &6     &1024   &85.4   &86.8 &93.5 &92.8 \\
            base model with 3Conv     &6        &6     &1024   &84.2   &86.6 &93.7 &92.8 \\
            base model with 7Conv     &6        &6     &1024   &77.4   &80.7 &89.1 &89.5 \\
            base model with 2CNNLSTM  &6        &6     &1024   &85.5   &86.1 &94.4 &94.9 \\
            \midrule
            big model with 2Conv      &6        &6     &1024   &86.5   &88.3 &95.4 &94.7 \\
            big model with 2CNNLSTM   &6        &6     &1024   &84.7   &84.8 &94.1 &93.4 \\
            \bottomrule
          \end{tabular}}
        \caption{\label{tab:table2} {\it Exploration of the encoder, the decoder and the modality-transform blocks on lexicon-free benchmarks.}}
\end{table}

\begin{figure}
\footnotesize
\centering
\includegraphics[width=5cm]{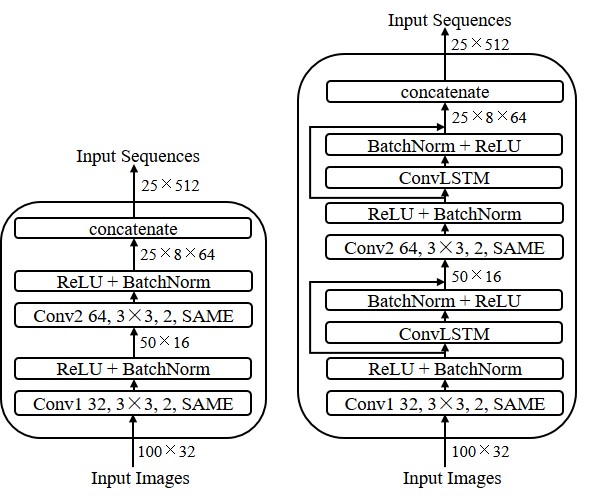}
\captionsetup{font={footnotesize}}
\caption{Examples of the proposed modality-transform block. (left) The general CNN block. (right) The CNNLSTM block.}
\label{fig3}
\end{figure}

\subsubsection{Exploration of the modality-transform block}
We investigate various architectures and depict two examples in Fig.\ref{fig3}.
As listed in Tab.\ref{tab:table2}, more convolutional layers (2Conv vs 3Conv, 7Conv) lead to a decline in performance, even a seven-layer convnet used in CRNN \cite{shi2017end} and RARE \cite{shi2016robust}.
We conjecture that the loss of detailed information due to resolution subsampling outweighs the gain of high-level semantics.
Since the encoder itself has strong feature extraction ability, we prefer to apply a two-layer convolution in the block and combine it with the encoder to draw more discriminative features.
Besides, as the newly CNNLSTM \cite{zhang2017very} could captures more temporal information by recurrent connections, we replace CNN with it. Results show an accuracy boost in our base model, but a little reduction in our big model.
The reason we guess is the redundant extraction of image information when associates CNNLSTM with excessive encoder components.

\subsection{Comparisons with the State-of-the-arts}
Based on the above analysis, we construct the final NRTR by setting ${N_e} = 12$, ${N_d} = 6$, ${d_{ff}} = 4096$ and the modality-transform block with two convolutional layers.
Since recent methods are trained with both Synth90k \cite{jaderberg2014synthetic} and SynthText \cite{gupta2016synthetic}, for fair comparison, we further train NRTR on two synthetic datasets.
Quantitative results are listed in Tab.\ref{tab:table1}.

\subsubsection{Accuracy}
For regular benchmarks, NRTR shows an accuracy boost and averagely beats previous best recognizers \cite{cheng2017focusing,bai2018edit} in both lexicon-free and lexicon-based cases.
Specifically, NRTR obtains 0.9\% on IIIT5K with ”1k” lexicon, 1.5\% on SVT and 0.2\% on IC03 with ”Full” lexicon, only litter decline on IIIT5K (0.3\%) with”50” lexicons.
In the lexicon-free case, NRTR surpasses \cite{bai2018edit} on all benchmarks.

Moreover, we test NRTR on irregular benchmarks and show results in Tab.\ref{tab:table3}.
Note that comparative models are all designed specially for irregular texts.
NRTR does not perform any special operation but still shows great tolerance on handling irregular texts, which further illustrates its strong ability in text feature extraction.

\begin{table}
\footnotesize
        \captionsetup{font={footnotesize}}
        \centering
          \scalebox{0.58}{
          \begin{tabular}{lcccccccccc}
            \toprule
            \multirow{3}{*}{Methods}    &\multicolumn{9}{c}{Regular Text}  &Training Time\\
            \cmidrule{2-11}
            &\multicolumn{3}{c}{IIIT5K}  &\multicolumn{2}{c}{SVT}  &\multicolumn{3}{c}{IC03}  &IC13   &\multirow{3}{*}{h/GPU}\\
            \cmidrule{2-9}    \cmidrule{10-10}
            &50 &1k &None     &50 &None         &50 &Full &None     &None                    \\
           \midrule
            ABBYY\cite{wang2011end}                           &24.3 &-    &-      &35.0 &-        &56.0 &55.0 &-       &-      &-\\
            Wang et al.\cite{wang2011end}                     &-    &-    &-      &57.0 &-        &76.0 &62.0 &-       &-      &-\\
            Mishra et al.\cite{mishra2012scene}               &64.1 &57.5 &-      &73.2 &-        &81.8 &67.8 &-       &-      &-\\
            Goel et al.\cite{goel2013whole}                   &-    &-    &-      &77.3 &-        &89.7 &-    &-       &-      &-\\
            Bissacco et al.\cite{bissacco2013photoocr}        &-    &-    &-      &90.4 &78.0     &-    &-    &-       &87.6   &-\\
            Alsharif et al.\cite{alsharif2013end}  &-    &-    &-      &74.3 &-        &93.1 &88.6 &-       &-      &-\\
            Alm\'azan et al.\cite{almazan2014word}            &91.2 &82.1 &-      &89.2 &-        &-    &-    &-       &-      &-\\
            Yao et al.\cite{yao2014strokelets}                &80.2 &69.3 &-      &75.9 &-        &88.5 &80.3 &-       &-      &-\\
            Jaderberg et al.\cite{jaderberg2014deep1}         &- &- &-            &86.1 &-        &96.2 &91.5 &-       &-      &-\\
            Su and Lu et al.\cite{su2014accurate}             &-    &-    &-      &83.0 &-        &92.0 &82.0 &-       &-      &-\\
            Jaderberg et al.\cite{jaderberg2016reading}       &97.1 &92.7 &-      &95.4 &80.7     &98.7 &\textbf{98.6} &93.1   &90.8     &-\\
            Lee et al.\cite{lee2016recursive}                 &96.8 &94.4 &-      &96.3 &80.7     &97.9 &97.0 &88.7    &90.0   &-\\
            Shi et al.\cite{shi2016robust}                    &96.2 &93.8 &81.9   &95.5 &81.9     &98.3 &96.2 &90.1    &88.6   &16/Titan X\\
            Shi et al.\cite{shi2017end}                       &97.6 &94.4 &78.2   &96.4 &80.8     &98.7 &97.6 &89.4    &86.7   &-\\
            Ghosh et al.\cite{ghosh2017visual}                &-    &-    &-      &95.2 &80.4     &95.7 &94.1 &92.6    &-      &-\\
            Yin et al.\cite{yin2017scene}                     &98.9 &96.7 &81.6   &95.1 &76.5     &97.7 &96.4 &84.5    &85.2   &-\\
            Gao et al.\cite{gao2017reading}                   &99.1 &97.9 &-      &97.4 &82.7     &98.7 &96.7 &89.2    &88.0   &-\\
            Cheng et al.$^*$\cite{cheng2017focusing}         &99.3 &97.5 &87.4   &97.1 &85.9     &\textbf{99.2} &97.3 &94.2    &93.3      &40/M40 \\
            Bai et al.$^*$\cite{bai2018edit}                 &\textbf{99.5} &97.9 &88.3   &96.6 &87.5     &98.7 &97.9 &94.6    &94.4   &41/P40 \\
            \midrule
            Our proposed NRTR                                 &99.2 &98.4 &86.5   &98.0 &88.3     &98.9 &97.9 &\textbf{95.4}    &94.7      &2.8/Titan X \\
            Our proposed NRTR$^*$                             &99.2 &\textbf{98.8} &\textbf{90.1}   &\textbf{98.1} &\textbf{91.5}     &98.9 &98.0 &94.7    &\textbf{95.8}  &5.0/Titan X \\
            \bottomrule
          \end{tabular}}
          \caption{\label{tab:table1} {\it Accuracies (\%) on regular benchmarks. "50", "1k" and "Full" are lexicon sizes. "None" means the lexicon-free case. 'h/GPU' indicates training time cost per epoch on their GPUs. Note that the FLOPS is $P40 > M40 \approx Titan X$. $^*$ means training with both Synth90k and SynthText.}}
\end{table}

\begin{table}
\footnotesize
        \captionsetup{font={footnotesize}}
        \centering
          \scalebox{0.55}{
          \begin{tabular}{lcccccc}
            \toprule
            \multirow{3}{*}{Methods}    &\multicolumn{6}{c}{Irregular Text} \\
            \cmidrule{2-7}
            &IC15  &\multirow{3}{*}{ } &\multicolumn{2}{c}{SVT-P} &\multirow{3}{*}{ } &CUTE80\\
            \cmidrule{2-7}
                                                     &None     & &50 &None     & &None        \\
            \midrule
            AON$^*$\cite{cheng2018aon}               &68.2     & &94.0 &73.0   & &76.8\\
            Aster$^*$\cite{shi2018aster}             &76.1     & &-    &78.5   & &79.5\\
            Liao et al.$^*$\cite{liao2018scene}      &-        & &- &-         & &79.9\\
            SAR\cite{li2018show}                     &78.8     & &\textbf{95.8} &86.4   & &\textbf{89.6}\\
            \midrule
            Our proposed NRTR$^*$                    &\textbf{79.4} & &94.9 &\textbf{86.6}   & &80.9\\
            \bottomrule
          \end{tabular}}
          \caption{\label{tab:table3} {\it Accuracies (\%) on irregular benchmarks. $^*$ means training with both Synth90k and SynthText.}}
\end{table}
\subsubsection{Speed} We give training speed of these approaches in Tab.\ref{tab:table1}. Only a few methods report their training time. For each epoch, Shi et al.\cite{shi2016robust} costs 16 hours/Titan X and Cheng et al.\cite{cheng2017focusing} needs 40 hours/Tesla M40. Both need 3 epochs before converging. The best previous method Bai et al.\cite{bai2018edit} takes more time (41 hours per epoch on P40). NRTR takes merely $5.0$ hours per epoch and therefore is at least 8 times faster than the existing best recognizer.
The inference speed of NRTR is approximately 0.03s per image, compared to 0.11s in \cite{bai2018edit} and 0.2s in \cite{shi2016robust}.

\subsubsection{Visualization} We show both correct and incorrect examples of NRTR in Fig.\ref{fig4}.
As can be seen, NRTR could recognize extremely challenging scene images, e.g., low resolution, complex geometric deformations and cluttered background. Some are even hard to human.
We carefully analyze incorrect results and split them into three types according to caused reasons.
First, texts are severely occluded by other objects, e.g., tree or barrier in example of 'redwood'.
Second, characters that look similar are mixed, like 'i' in image of 'valerie' and its fault result 'l'.
Third, text orientation are seriously curved, e.g, nearly ninety degrees to the horizontal plane.
These failed examples also highlight future research directions of the proposed NRTR.

\section{Conclusions}
This paper points out two problems lying in current RNN/CNN-based scene text recognizers and proposes a no-recurrence model aiming at increasing computation parallelization and performance. Experiments demonstrate its superiority on accuracy and training speed. We intend to extend the idea to end-to-end text spotting system.

\section*{Acknowledgment}
This work is supported by the Strategic Priority Research Program of the Chinese Academy of Sciences (XDBS01070101).



%
%
%

\bibliographystyle{IEEEtran}
\bibliography{refs}

\end{document}